\documentclass[preprint,12pt]{elsarticle}

\usepackage{amssymb}
\usepackage{multirow}
\usepackage{amsmath}

\journal{Nuclear Physics B}

\begin{document}

\begin{frontmatter}



\title{Image-Plane Geometric Decoding for View-Invariant Indoor Scene Reconstruction}


\author[1]{Mingyang Li}
\ead{limingyang97@tju.edu.cn}
\affiliation[1]{organization={School of Microelectronics,Tianjin University},
    city={Tianjin},
    postcode={300072}, 
        country={China}}
\fnmark[fn1]
\author[1]{Yimeng Fan}
\ead{yimengfan@tju.edu.cn}

\author[1]{Changsong Liu}
\ead{changsong@tju.edu.cn}

\author[1]{Lixue Xu}
\ead{xlx_13@tju.edu.cn}

\author[1,3]{Xin Wang}
\ead{wangxin@tfri.com.cn}

\author[2]{Yanyan Liu \corref{cor1}}
\ead{lyytianjin@nankai.edu.cn}

\author[1]{Wei Zhang \corref{cor1}}

\ead{tjuzhangwei@tju.edu.cn}



\affiliation[2]{organization={College of Electronic Information and Optical Engineering,Nankai University}, 
    city={Tianjin},
    postcode={300072}, 
        country={China}}

\affiliation[3]{organization={Tianjin Fire Science and Technology Research Institute of MEM}, 
    city={Tianjin},
    postcode={300381}, 
        country={China}}

\cortext[cor1]{Corresponding author}

\fntext[fn1]{This is the first author footnote.}

\begin{abstract}
Volume-based indoor scene reconstruction methods offer superior generalization capability and real-time deployment potential. However, existing methods rely on multi-view pixel back-projection ray intersections as weak geometric constraints to determine spatial positions. This dependence results in reconstruction quality being heavily influenced by input view density. Performance degrades in overlapping regions and unobserved areas.To address these limitations, we reduce dependency on inter-view geometric constraints by exploiting spatial information within individual views. We propose an image-plane decoding framework with three core components: Pixel-level Confidence Encoder, Affine Compensation Module, and Image-Plane Spatial Decoder. These modules decode three-dimensional structural information encoded in images through physical imaging processes. The framework effectively preserves spatial geometric features including edges, hollow structures, and complex textures. It significantly enhances view-invariant reconstruction.Experiments on indoor scene reconstruction datasets confirm superior reconstruction stability. Our method maintains nearly identical quality when view count reduces by 40\%. It achieves a coefficient of variation of 0.24\%, performance retention rate of 99.7\%, and maximum performance drop of 0.42\%. These results demonstrate that exploiting intra-view spatial information provides a robust solution for view-limited scenarios in practical applications.
\end{abstract}

\begin{graphicalabstract}
\includegraphics[width=\textwidth]{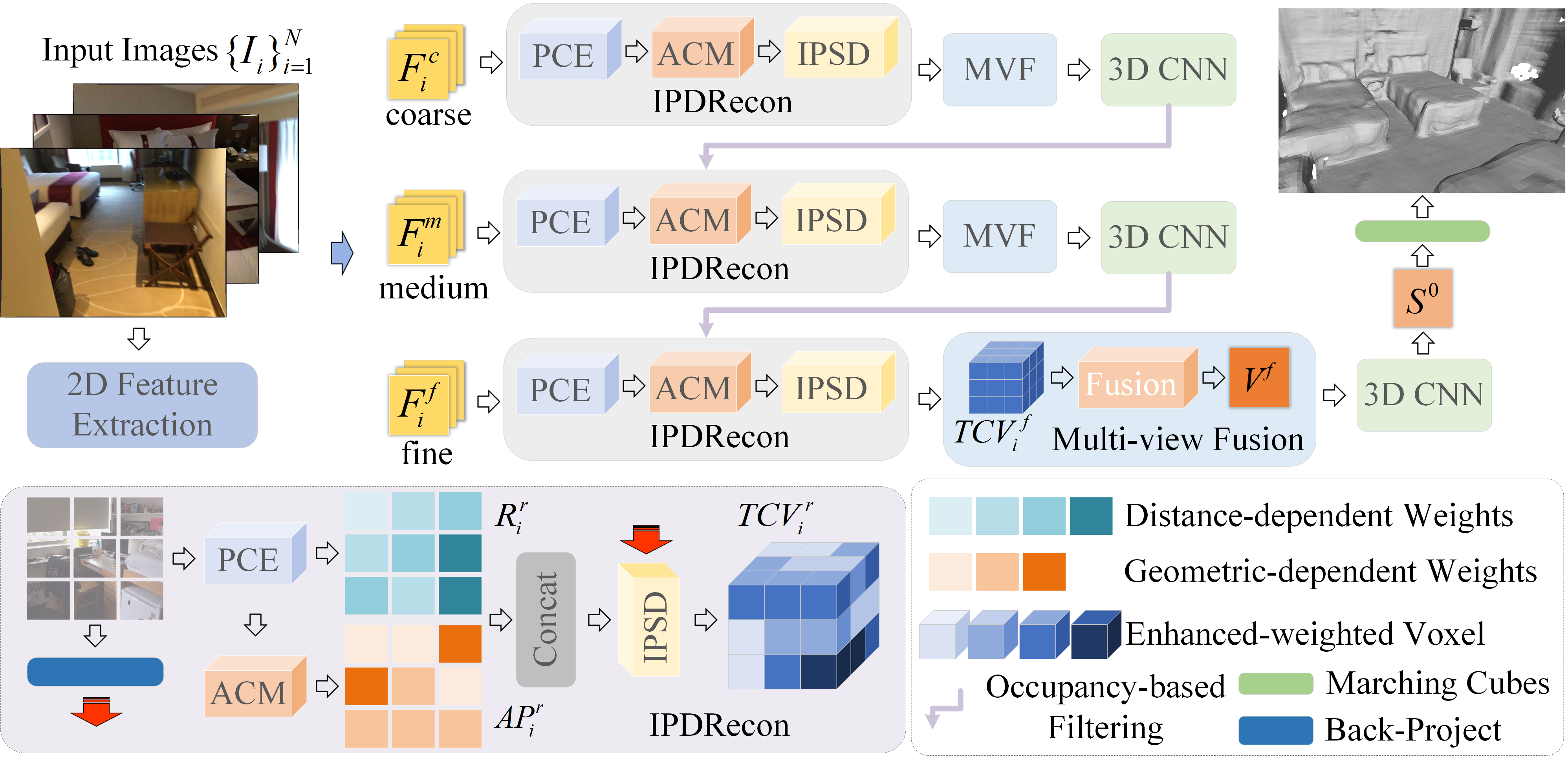}
\end{graphicalabstract}

\begin{highlights}
\item Novel image-plane decoding framework exploits intra-view geometric constraints, reducing dependency on multi-view intersections for reconstruction.
\item Pixel-level Confidence Encoder and Affine Compensation Module extract distance, position, and geometric features from single views.
\item Achieves 99.7\% performance retention with 40\% fewer views, demonstrating superior view-invariant reconstruction stability.
\end{highlights}


\end{frontmatter}



\section{Introduction}
With the growing demand for intelligent living, the requirements for autonomous perception capabilities of devices are continuously escalating. Traditional two-dimensional (2D) vision, as the primary means for devices to understand their environment, can no longer satisfy functional requirements \citep{surfacenet}. Consequently, three-dimensional (3D) vision has garnered increasing attention. Among these technologies, 3D reconstruction has emerged as a core technology in 3D vision, undergoing rapid iterative development \citep{yang2024sa}. Indoor scene reconstruction, in particular, has demonstrated remarkable vitality due to its extensive application prospects in smart homes, indoor surveillance, and virtual reality (VR)/augmented reality (AR) domains \citep{jiang2024geometry}.

Early indoor scene reconstruction methods combined traditional visual cameras with LiDAR systems. Typical approaches include Simultaneous Localization and Mapping (SLAM) \citep{spslam}  and Structure from Motion (SFM) \citep{sfm,10148998}. These methods face performance constraints from depth sensing hardware (LiDAR and depth cameras) and post-processing platforms. They fail to handle non-Lambertian surfaces, low-texture regions, and specular reflective surfaces \citep{10126077}. This results in incomplete reconstructions. Deep learning advances have introduced end-to-end approaches that eliminate depth input requirements. Among these, volumetric representation-based reconstruction methods maintain strong research prospects due to their robust generalization capabilities and real-time deployment potential \citep{neuralrecon,atlas}.

\begin{figure*}[!t]
\centering
\includegraphics[width=\textwidth]{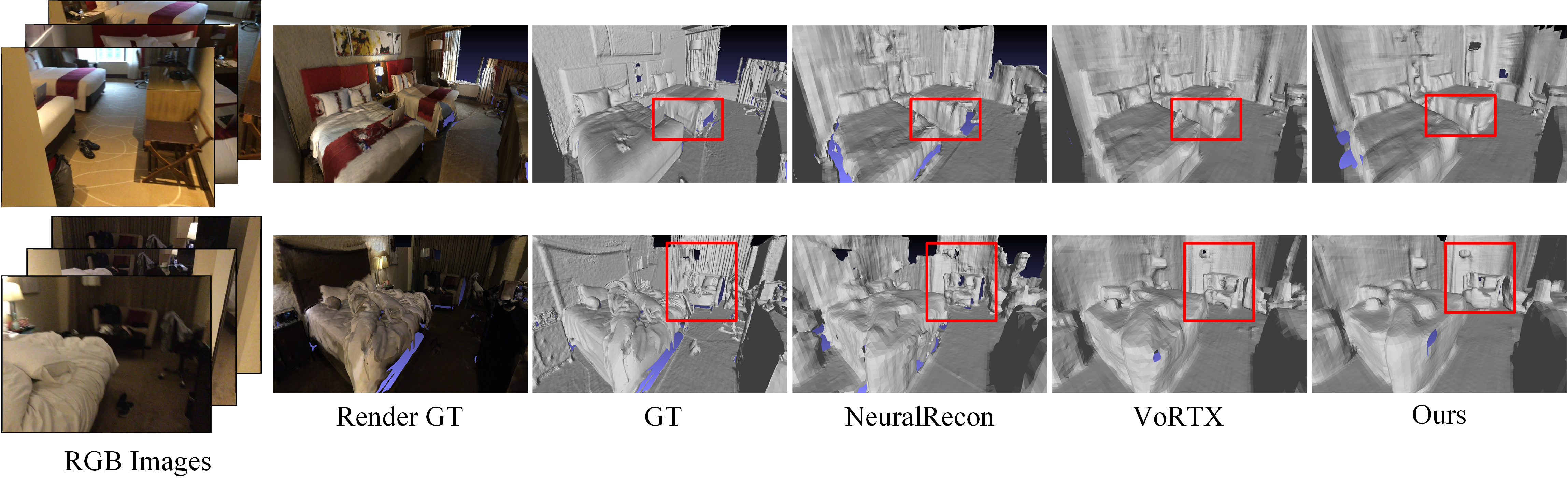}
\caption{Visualization of Reconstruction Quality. Previous methods suffer from severe artifacts and incomplete structures under sparse views. By leveraging single-view geometric priors through PCE and ACM modules, IPDRecon achieves superior reconstruction quality with better structural detail preservation.}
\label{fig_1}
\end{figure*}

Typical volumetric frameworks comprise three stages: 2D feature extraction, back-projection, and surface regression, taking registered RGB images as input to generate reconstructed scenes represented as signed distance fields (SDF). However, the fundamental limitation of existing frameworks lies in their back-projection process, which relies entirely on weak geometric constraints from multi-view ray intersection to determine spatial positions, rendering reconstruction quality heavily dependent on the number of views and angular differences between adjacent viewpoints \citep{chen2023deep,chen2025high}. As demonstrated in Table \ref{table1}, our validation experiments on ScanNetV2 confirm that reconstruction quality improves significantly with increasing view counts, underscoring the fundamental limitations of current multi-view dependent approaches. To mitigate view count constraints, existing methods have attempted to improve the back-projection process through strategies such as view weighting \citep{vortx}, auxiliary frames \citep{cvrecon}, and additional occupancy supervision \citep{inner}. However, as illustrated in Figure \ref{fig_1}, these approaches continue to exhibit artifacts including missing sharp edges, noise accumulation in concave structures, and excessive smoothing in complex regions. The core issue remains that limited views (60-100 images) in large-scale indoor scenes cannot provide sufficient geometric constraints. Several large-scale vision models \citep{dpa,wang2025vggt} have attempted to reduce dependence on view counts in 3D perception by learning pixel feature correlations within image planes through large-scale data pre-training to predict distance information. Nevertheless, such purely data-driven approaches still exhibit generalization limitations in complex indoor scenes with insufficient training data and cannot be directly integrated into geometry-constraint-based back-projection frameworks.

\begin{table}[t]
\centering
\caption{Experimental results of view number variations on the ScanNet V2 dataset \citep{scannet}. VoRTX \citep{vortx} was selected as the test model. The tested view numbers were 60, 80, and 100. Due to computational resource limitations, the maximum number of views was set to 100. The reconstruction quality (precision, recall, and F-score) increases significantly as the number of views grows. This validates the dependency of back projection effectiveness on view number configuration.}
\begin{tabular}{c|ccccc}
\hline
Methods & Acc $\downarrow$ & Comp $\downarrow$ & Prec $\uparrow$ & Recall $\uparrow$ & F-score $\uparrow$ \\
\hline
60 views & 0.046 & 0.075 & 0.752 & 0.631 & 0.685 \\
80 views  & 0.043 & 0.075 & 0.763 & 0.639 & 0.694 \\
100 views & 0.043 & 0.072 & 0.767 & 0.651 & 0.703 \\
\hline
\end{tabular}
\label{table1}
\end{table}

Inspired by the idea of extracting intra-view correlations, we propose an Image-Plane Decoding framework for indoor 3D reconstruction (IPDRecon). Unlike existing Vision Large Models (VLMs), we propose leveraging the physical principles of geometric optical imaging as inductive bias. This approach systematically exploits rich geometric constraints encoded within single views. It eliminates dependence on large-scale data and endows models with intra-view 3D information perception capabilities. This fundamentally reduces reliance on view count and establishes stronger geometric constraints. IPDRecon comprises three core components: Pixel-level Confidence Encoder (PCE), Affine Compensation Module (ACM), and Image-Plane Spatial Decoder (IPSD). First, we propose PCE to endow the model with preliminary 3D information perception capability. PCE integrates state-space modeling mechanisms with diffuse reflection imaging geometry properties \citep{lambert}, enabling the model to acquire initial distance and position awareness. Since preliminary 3D information struggles to maintain complex geometric structures within space, we introduce ACM to provide rigid geometric constraints. We observe that different affine transformations preserve distinct properties (affine invariance) \citep{affin}. ACM constructs local affine-invariant features to impose 3D rigid geometric constraints, ensuring geometric accuracy of complex structures under sparse views. IPSD presents a novel multi-source geometric prior fusion strategy. It jointly encodes distance, position, and geometric features within the cost volume, fundamentally improving traditional back-projection representation capability. By systematically leveraging single-view geometric information to strengthen intra-view geometric constraints, IPDRecon addresses the dependency limitations of existing frameworks on keyframe quality and quantity. Experimental results on ScanNet V2 demonstrate that IPDRecon significantly reduces reconstruction artifacts in sparse viewpoint scenarios while enhancing reconstruction stability.

Our main contributions are as follows:

(1) We propose IPDRecon, a novel Image-Plane Decoding framework that systematically exploits single-view geometric information to strengthen intra-view geometric constraints. By integrating state-space modeling with diffuse reflection imaging principles in PCE and leveraging affine invariance for rigid geometric constraints in ACM, our framework effectively reduces dependency on view quantity in indoor 3D reconstruction.

(2) We design an innovative multi-source geometric prior fusion strategy in IPSD that jointly encodes distance, position, and geometric features within the cost volume. This approach fundamentally enhances traditional back-projection representation capability, enabling high-fidelity reconstruction under sparse viewpoint conditions while preserving complex structural details.

(3) Extensive experiments on ScanNet V2 demonstrate IPDRecon's effectiveness, achieving $79.7\%$ Precision and $0.722$ F-score. Our method significantly reduces reconstruction artifacts in sparse viewpoint scenarios while achieving superior view-invariance, providing a practical solution for real-world indoor scene reconstruction.

The paper is organized as follows: Section 2 surveys recent methodologies in indoor 3D reconstruction. Section 3 elaborates on the implementation details of IPDRecon. Section 4 presents our experimental protocols and results analysis. Section 5 concludes with a summary and prospects for future work.

\section{Related Work}
Indoor scene reconstruction has garnered significant attention due to its broad application demands. Researchers have constructed a three-stage framework. Current methods are primarily categorized into depth-based, volume-based, and neural implicit representation methods.

\subsection{Depth-based Methods}

Depth-based 3D reconstruction typically utilizes multi-view depth information to achieve 3D structure reconstruction. Traditional methods \citep{15multi,pixelwise,patchmatch} estimate depth through photometric and geometric consistency, followed by weighted fusion of depth information. Learning-based methods \citep{mvsnet,deepvideomvs,kinectfusion,neuralfusion,routedfusion,simplerecon} have gradually gained attention. MVSNet \citep{mvsnet} constructed cost volumes from associated frames to estimate accurate depth information. Some researchers introduced external information such as Gaussian priors \citep{dp}, Bayesian theory \citep{neuralrgbd}, and optical flow \citep{videof} for depth map prediction. DVS \citep{deepvideomvs} used LSTM to predict depth. SimpleRecon \citep{simplerecon} constructed cost volumes using camera metadata, avoiding 3D convolution. FineRecon \citep{finerecon} combined depth prediction with TSDF occupancy prediction, using 3D-CNN to process information from both branches to generate fine surfaces.

\subsection{Volume-based Methods}

To overcome depth information reliance, researchers have explored volume-based methods for end-to-end 3D reconstruction. \citep{surfacenet} is a pioneering work. Atlas \citep{atlas} and NeuralRecon \citep{neuralrecon} introduced a three-stage coarse-to-fine pipeline, extending the task to multi-view sequences. Some researchers \citep{transformerfusion,vortx} incorporated transformer architectures in the view fusion stage, replacing the traditional averaging methods. CVRecon \citep{cvrecon} constructed cost volumes from associated frames to suppress noise from single-ray projections. VisFusion \citep{visfusion} improved feature fusion by explicitly inferring voxel visibility. IOAR \citep{inner} added intra-object awareness networks to the coarse and medium stages to intervene in the regression process of internal and external object surfaces. SDFFormer \citep{SDFFORMER} provided a novel backbone for coarse-to-fine reconstruction tasks. SDFUtasn \citep{li2025hybrid} proposed a novel 3D backbone that utilizes sparse convolution characteristics to maintain surfaces with enhanced spatial details during the final prediction stage. Despite compensating with additional information, volume-based methods face theoretical limitations in back-projection processes, making fine surface reconstruction challenging and resource-intensive.

\subsection{Neural Implicit Representation Methods}
Neural implicit representations use neural networks to learn intermediate feature states. The introduction of neural radiance fields \citep{nerf} has advanced this approach. Early research focused on addressing occlusion and optimizing convergence efficiency \citep{pixelnerf,mipnerf,mine}, but large-scale scene reconstruction remains challenging \citep{wen2025nerf}. Some studies \citep{nerfusion,pointnerf} achieved scalable NeRF representations for multi-view feature aggregation but primarily focused on novel view synthesis, lacking high-quality surface reconstruction. NeuS \citep{neus} extended NeRF by incorporating volume rendering techniques. Other works \citep{manhattan,neurmips} introduced fundamental reconstruction principles and segmentation priors to optimize quality. Subsequent research incorporated priors like color \citep{nfs}, depth \citep{scenerf}, and primitives \citep{mobilenerf} to supervise implicit learning processes. Additionally, some studies \citep{zhang2022critical,fu2022geo} introduced explicit representation priors and hyperplane supervision to mitigate the impact of large-scale dataset training. While neural implicit representation-based reconstruction can predict high-quality surfaces, issues such as limited scalability, high resource consumption, and lengthy rendering times hinder real-time applications.

Rather than solely relying on idealized multi-view ray intersections as weak geometric constraints, we exploit geometric optical imaging principles to systematically mine rich three-dimensional information within individual views. We propose an Image-Plane Decoding framework for indoor scene reconstruction (IPDRecon) that effectively integrates single-view geometric priors with multi-view geometric constraints. Through Pixel-level Confidence Encoder (PCE) and Affine Compensation Module (ACM), we systematically extract intra-view geometric relationships and encode spatial distance deviations as pixel-level confidence weights. Subsequently, our Image-Plane Spatial Decoder (IPSD) integrates these intra-view geometric priors with multi-view constraints, strengthening both intra-view and inter-view geometric constraints to achieve robust 3D reconstruction with enhanced spatial detail fidelity.

\begin{figure*}[t]
\centering
\includegraphics[width=\textwidth]{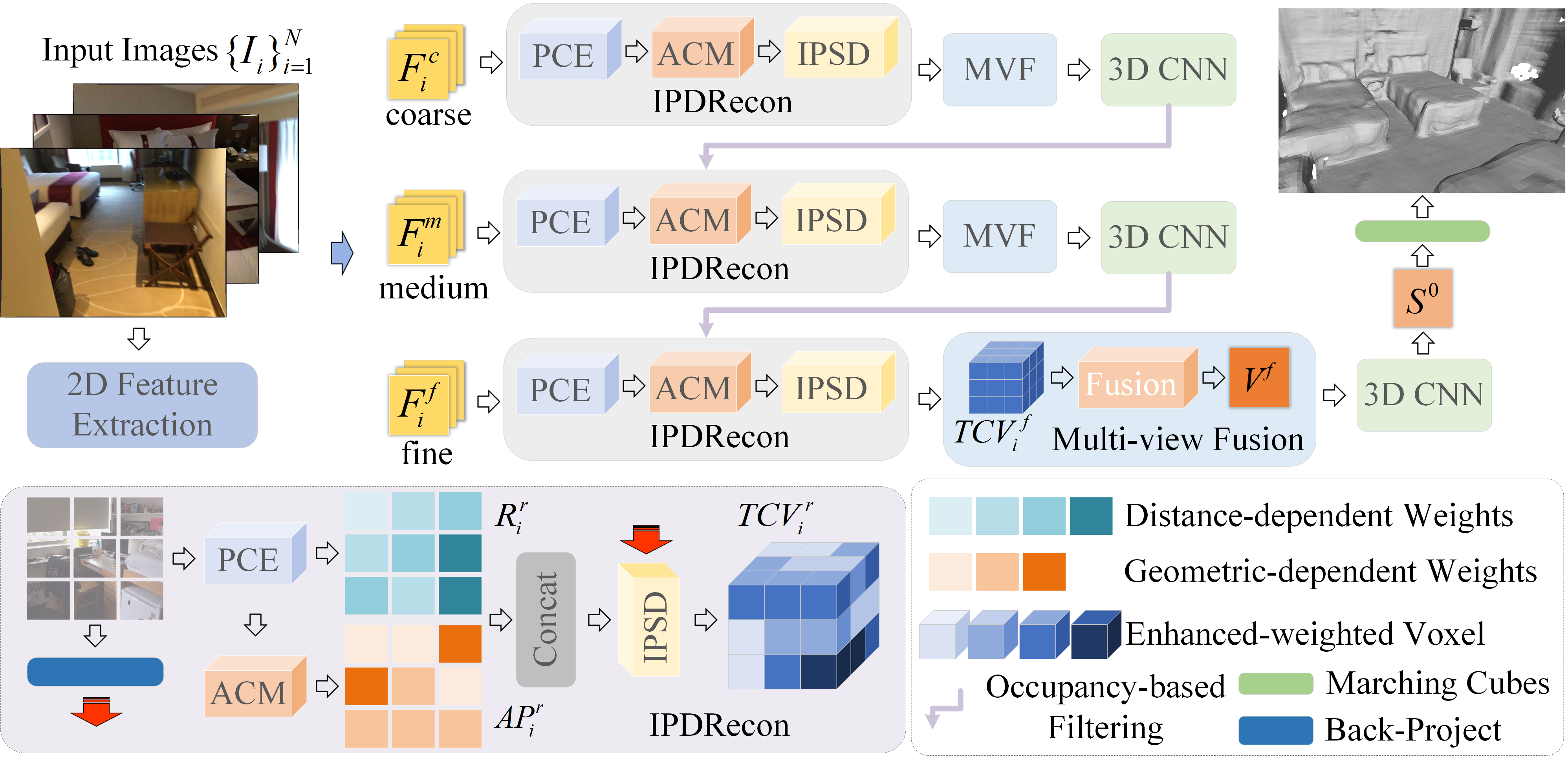} 
\caption{Architecture of IPDRecon. Given a series of posed images, we use a 2D backbone network to generate coarse-to-fine 2D features $F_{i}^{r}$. Subsequently, through the IPD-Projection stage and the multi-view fusion stage, we obtain the feature volume $V^{r}$. Finally, we use a 3D backbone network to regress the scene surface.}
\label{fig2}
\end{figure*}

\section{Methodology}

In this section, we detail the implementation specifics of our proposed Image-Plane Decoding framework (IPDRecon). Section 3.1 details the reconstruction pipeline of IPDRecon. Section 3.2 elaborates on the implementation details of the Pixel-level Confidence Encoder (PCE). Section 3.3 describes the primary workflow of the Affine Compensation Module (ACM). Section 3.4 explains the construction methodology of the Image-plane Spatial Decoder (IPSD). Section 3.5 outlines the supervision approach for IPDRecon.

\subsection{Overview}

As shown in Figure \ref{fig2}, IPDRecon architecture adopts a coarse-to-fine pipeline. For the image sequence $\left\{I_{i}\right\}_{i=1}^{N}$ and camera extrinsic parameters $\left\{K_{i}, P_{i}\right\}_{i=1}^{N}$ after keyframe selection \citep{vortx}, we use the MnasNet-B1 \citep{mnasnet} backbone network to extract features and perform bottom-up multi-scale fusion to obtain multi-resolution 2D features $\left\{F_{i}^{r}\right\}_{i}^{N}$, here $ r \in\{ coarse,medium,fine \}$. For simplicity, we focus on the features $F_{i}^{r}$ for a particular frame at specific resolution.

During IPDRecon stage, the Pixel-level Confidence Encoder (PCE) first encodes $F_{i}^{r}$ to generate ray-encoded features $R_{i}^{r}$ with per-pixel weights. Subsequently, the Affine Compensation Modules (ACM) generates the geometric-enhanced feature sequence $A P_{i}^{r}$. Finally, the Image-plane Spatial Decoder (IPSD) captures the differences in ray contributions caused by affine projections $A P_{i}^{r}$ and confidence-weighted feature $R_{i}^{r}$ to construct the 3D cost volume $TCV_{i}^{r}$.

Since $TCV_{i}^{r}$ contains information about a single view , we need to fuse the cost volumes of all selected keyframes. However, a simple average fusion cannot differentially preserve the spatial and geometric information of each view. Therefore, we use the Transformer $t^{r}(\cdot)$ to assign fusion weights $W_{i}^{r}$ to each view to construct the fused feature volume $V^{r}=\left\{v_{1}^{r}, v_{2}^{r}, \cdots, v_{N}^{r}\right\}$:

\begin{equation}
V^{r} = \left\{v_{1}^{r}, v_{2}^{r}, \cdots, v_{N}^{r}\right\}=t^{r}\left(\left\{T C V_{i}^{r}\right\}_{i=1}^{N}\right).
\end{equation}

In the fine stage, the network predicts the TSDF volume $S^{0}$ and extracts the zero-level surface using Marching Cubes \citep{marchingCUBE}. At coarse and medium stages, the network employs independent occupancy prediction branches. These branches filter noisy occupancy from coarse to fine for subsequent stages (as indicated by the purple arrows in Figure \ref{fig2}).

\subsection{Pixel-level Confidence Encoder (PCE)}

\begin{figure}
	\centering
		\includegraphics[width=3.5in]{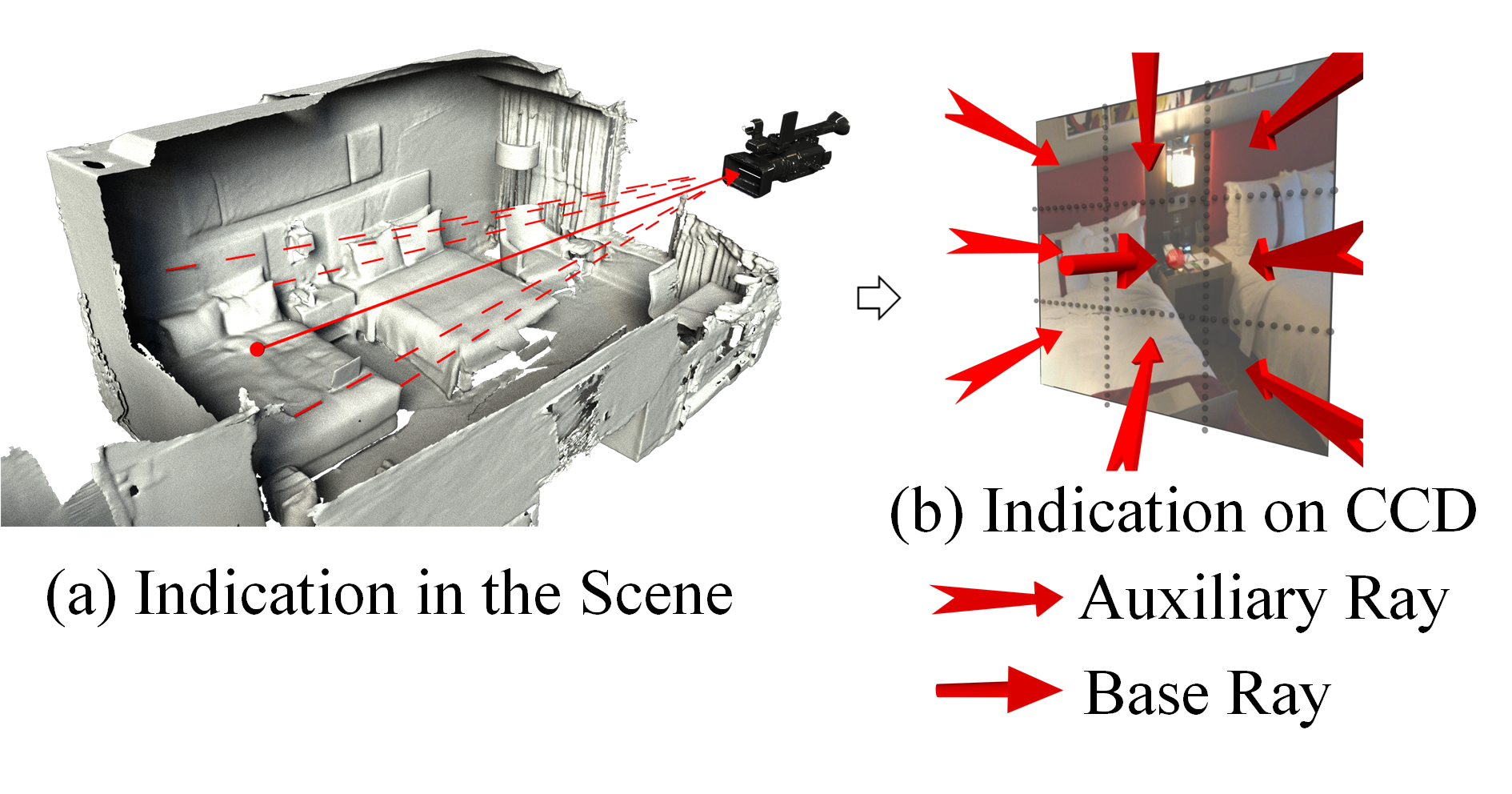}
	\caption{According to Lambert's law, the interaction of reflected light within a scene contributes to imaging. Consequently, the information of a single pixel is formed by the linear superposition of multiple light rays from the spatial domain.}
	\label{FIG:3}
\end{figure}

Existing methods rely solely on multi-view ray intersections as weak geometric constraints to determine 3D spatial point occupancy. This causes reconstruction quality to depend heavily on keyframe selection and quantity. The occupancy probability of 3D points becomes proportional to the number of intersecting back-projection rays at each location. Recent large vision models \citep{dpa,dpa2,wang2025vggt} leverage Transformer frameworks' global modeling capability to mine intra-view and inter-view geometric correlations for 3D perception. While these approaches reduce dependency on view quantity, they require extensive pre-training on large-scale datasets and exhibit limited generalization to complex indoor scenes with insufficient training data. Moreover, these purely data-driven methods cannot be directly integrated into geometry-constraint-based back-projection frameworks due to their fundamentally different architectural paradigms.

Inspired by the concept of extracting intra-view correlations but addressing the limitations of data-driven approaches, we propose a Pixel-level Confidence Encoder (PCE) to endow the model with preliminary 3D information perception capability. Unlike large vision models that enhance perception through guided learning, we integrate state-space modeling mechanisms with diffuse reflection imaging geometry properties. This constructs an encoding architecture with inherent distance and position awareness. The design enables 2D-to-3D spatial reasoning at the architectural level, establishing a foundation for subsequent geometric constraint enforcement. Through PCE, we systematically extract intra-view geometric information, effectively reducing reconstruction dependency on view quantity. Section 3.2.1 details diffuse reflection-based imaging principles. Section 3.2.2 describes the PCE design.

\subsubsection{Mathematical Expressions for Diffuse Reflection Imaging}

As shown in Figure \ref{FIG:3}, light emitted from a source illuminates an object surface, reflects, and then converges onto a photosensitive element in the real world. This process adheres to Lambert's Law \citep{lambert}. Consequently, each pixel on the photosensitive element receives light that is, in fact, an aggregation of diffusely reflected rays from the entire scene. We refer to this aggregate as a light cluster. According to geometric optics, the process of imaging formed by the superposition of rays within a light cluster is linear. This process can be expressed as follows:

\begin{equation}
F_{I}(p)=\sum_{k \in {N}^{X \times Y \times Z}} \exp \left(-a d_{k}\right) \cdot F_{L}\left(P_{k}\right).
\end{equation}

For the feature $F_{I}(p)$ at point $p \in {R}^{U \times V}$ on the image plane after light cluster aggregation, it can be expressed as the sum of the contributions of light features $F_{L}\left(P_{k}\right)$ from various points in the scene, with $P_{k} \in {R}^{X \times Y \times Z}$. Since the superposition of light is influenced by factors such as distance and angle, which exhibit exponential attenuation or growth, the weight is denoted as $\exp \left(-a d_{k}\right)$, where $d_{k}$ represents the distance of the $k$-th ray, and $a$ is the attenuation coefficient. The diffuse reflection imaging process in Eq.(2) demonstrates that each pixel aggregates information from multiple spatial points with distance-dependent weights. This spatial aggregation pattern motivates the design of PCE to explicitly model pixel-level correlations for enhanced 3D perception.

\begin{figure}[t]
\centering
\includegraphics[width=9cm]{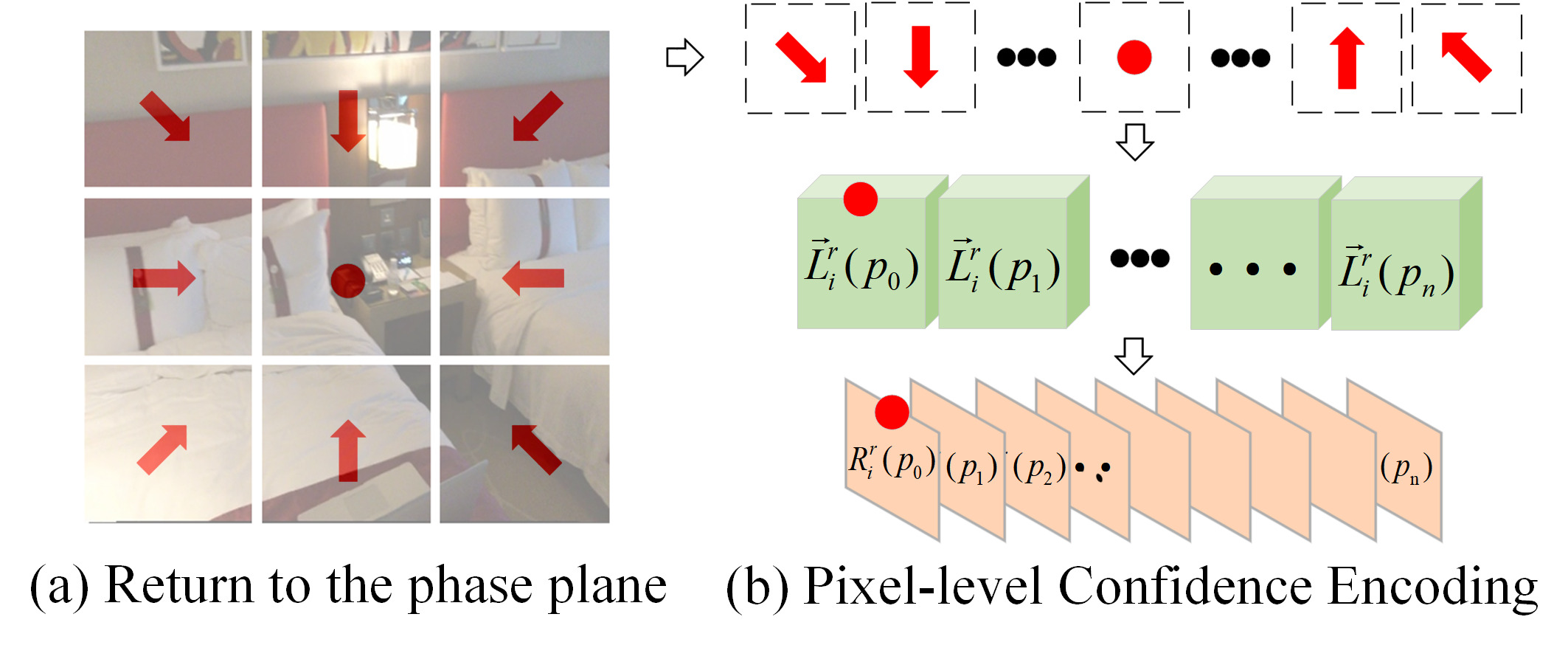} 
\caption{Demonstration of Pixel-level Confidence Encoder (PCE)  calculation process.}
\label{fig4}
\end{figure}

\subsubsection{Details of PCE}

Based on spatial aggregation principles observed in diffuse reflection imaging, PCE introduces state-space modeling for geometric information extraction in indoor scenes. Unlike direct application of generic state-space models such as Mamba \citep{mamba,vmunet,groupmamba}, we adapt the framework specifically for imaging characteristics: (1) mapping the distance attenuation factor exp(-adk) from Eq.(2) to confidence weight parameters in state space; (2) utilizing recursive properties of state space to model accumulation effects in light ray superposition. This domain-specific state-space modeling enables PCE to extract 3D geometric information embedded in 2D image planes, establishing the foundation for subsequent intra-view constraint enhancement.

To accommodate field-of-view limitations in 2D imagery, we impose a physical constraint on PCE: spatial points making significant contributions to the imaging of specific image plane points manifest on the feature plane due to distance constraints. According to this constraint, we transform the LPM encoding process into the confidence of all pixels $p_{j} \in {R}^{U \times V}$ on the feature plane for a specific position $p \in {R}^{U \times V}$.

As shown in Figure \ref{fig4}, we utilize the encoder to extract the correlations within the light cluster, fitting the superposition process of light in real space. For the light cluster mapping $\vec{L}_{i}^{r}(p)$, there is a linear relationship with the light cluster mapping $\vec{L}_{i}^{r}(p_{j})$. The spatial feature $F_{i}^{r}(p)$ can be understood as a mapping of  reflected light (primary projection rays). Therefore, $\vec{L}_{i}^{r}(p)$ can be expressed as:

\begin{equation}
\vec{L}_{i}^{r}(p)=\widehat{w} \cdot F_{i}^{r}(p)+\sum_{j \in {N}^{U \times V}} \hat{q}_{j} \vec{L}_{i}^{r}\left(p_{j}\right).
\end{equation}

Where, $\hat{q}_{j}$ represents the contribution weight of the light cluster mapping $\vec{L}_{i}^{r}(p_{j})$ to $\vec{L}_{i}^{r}(p)$, $\widehat{w}$ denotes the primary projection weight. Subsequently, we introduce distance information $d$ to discretize the space to obtain the final confidence weight $\widehat{q}_{j}=e^{du}$. In order to prevent the problem of vanishing or exploding gradient when the parameter updates, the initialization of the parameters follows the HiPPO theory \citep{mamba}, with the initialized parameters $u \in {R}^{M \times M}$ being sustainable and updatable. Additionally, the primary projection weight can be expressed as $\widehat{w}=\left(e^{du}-I\right) u^{-1} w$, where $w \in {R}^{C \times M}$ is a learnable parameter matrix randomly initialized following Gaussian distribution. Finally, we add confidence weights to the light cluster mapping $\vec{L}_{i}^{r}(p)$, obtaining light cluster encoded features ${R}_{i}^{r}(p)$ with per-pixel weights:

\begin{equation}
R_{i}^{r}(p)=MLP_{E}^{r}\left(L_{c}(p)\right).
\end{equation}

We use this weight to filter out light clusters on the feature plane that have the least correlation with other light rays. As the weight approaches zero, it indicates that the spatial point $P \in {R}^{X \times Y \times Z}$ corresponding to pixel $p_{j} \in {R}^{U \times V}$ is likely unoccupied in the space. Furthermore, the encoding process employs a linear attention mechanism that embeds distance relations $d$ and positional relations $u$ within the weights. Consequently, PCE can mine distance and positional information encoded within the 2D image plane by the diffuse reflection imaging process, endowing the network with 2D-to-3D perception capability and thereby reducing dependency on additional keyframes.

\subsection{Affine Compensation Modules (ACM)}

\begin{figure}[t]
\centering
\includegraphics[width=9cm]{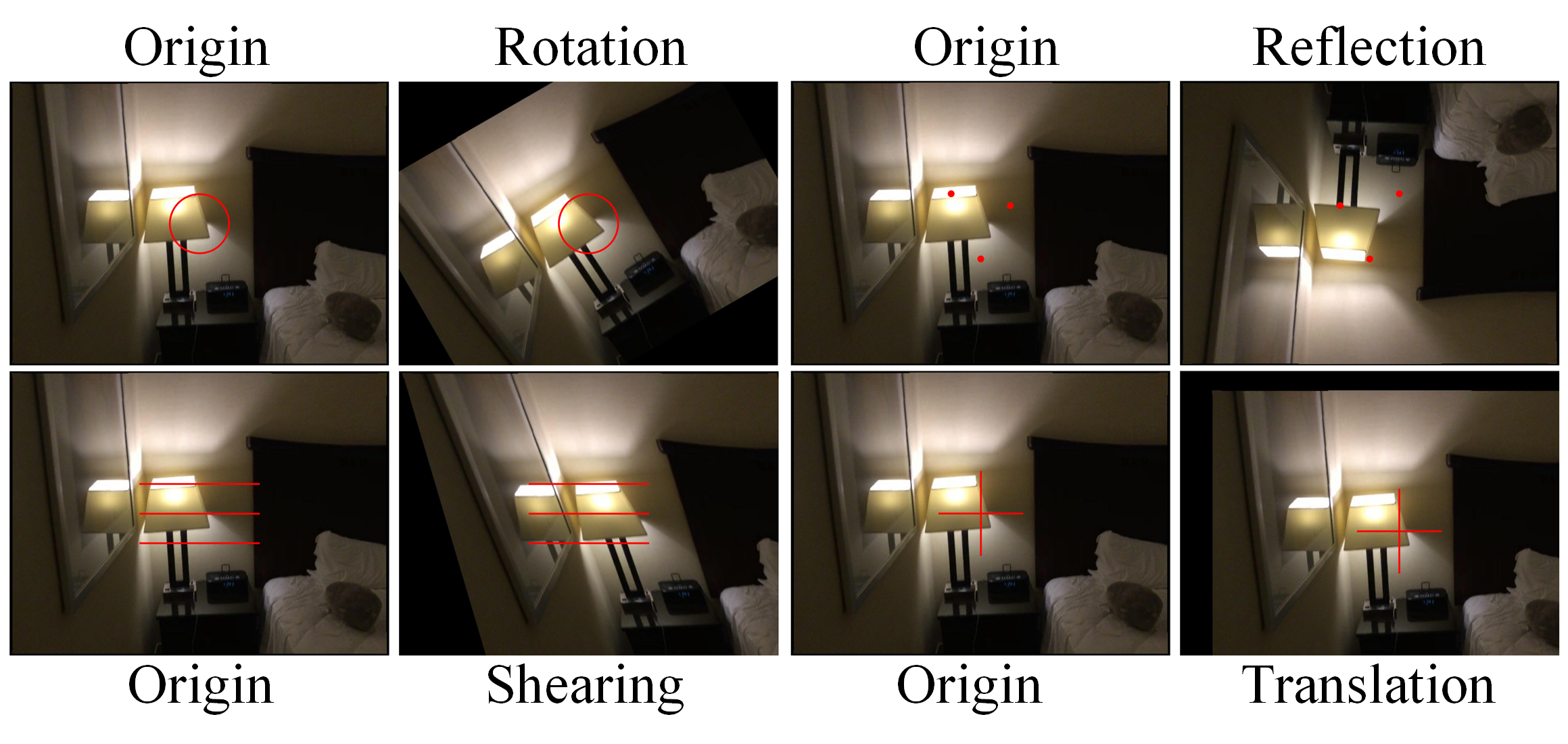} 
\caption{Visual representation of four affine transformations. Rotation and reflection alter the direction of light propagation while preserving path length and angular relationships between rays. Shearing transformation modifies the angular relationships between light rays while maintaining propagation direction and path length. Translation preserves all three properties: direction of propagation, path length, and angular relationships between light rays.}
\label{fig5}
\end{figure}

With PCE, the features $R_{i}^{r}$ contain preliminary geometric information. However, this information is maintained solely through spatial correlation weights on the feature plane. Without applying any geometric or optical constraints, this method struggles to represent complex structures within the scene. To decode the spatial information contained in the encoded features and to maintain geometric structure, we designed the Affine Compensation Modules (ACM). 

ACM employs affine invariance principles to enforce 3D rigid geometric constraints via local affine-invariant feature construction. As shown in Figure \ref{fig5}, we identify that distinct affine transformations preserve different geometric properties of reflected light rays: affine transformations excluding scaling maintain light path length; rotations and reflections modify light propagation direction whereas translations and scaling preserve it; shearing alters angular relationships between rays while other transformations maintain them \citep{affin}. Based on these Affine Invariance Mechanisms (AIM), ACM utilizes the line-preserving, parallelism-preserving, and ratio-preserving properties of affine transformations to extract affine-invariant geometric descriptors.

The affine transformation matrix $A_i^r$ is initialized as an identity matrix. To apply the affine transformation to input features $R_i^r(p)$, we first generate a sampling grid $G \in \mathbb{R}^{N \times H \times W \times 2}$ using normalized coordinates. The grid $G$ defines the spatial locations for resampling after affine transformation. We introduce channel-wise attenuation factors $b_c$ to modulate feature channels based on their geometric significance. The affine-compensated features for input $p(u,v)$ can be expressed as:

\begin{equation}
AP_i^r(\hat{p}) = F_{bilinear}(R_i^r, A_i^r \cdot G) \odot b_c.
\end{equation}

Here, $F_{bilinear}(\cdot)$ denotes bilinear interpolation for spatial transformation, and $\odot$ represents element-wise multiplication. The learnable affine transformation matrix $A_i^r$ captures the geometric relationship between 2D image features and 3D structural properties. Channel-wise attenuation factors $b_{c}$ selectively weight feature channels based on their geometric significance. The resulting affine-compensated features $AP_{i}^{r}(\hat{p})$ preserve affine-invariant geometric properties across viewpoints. We concatenate the original features $R_{i}^{r}(p)$ with $AP_{i}^{r}(\hat{p})$ to generate the enhanced geometric representation $CR_{i}^{r}(p)=\{R_{i}^{r}(p), AP_{i}^{r}(\hat{p})\}$. This dual representation ensures geometric consistency across different viewpoints, effectively providing rigid geometric constraints for 3D reconstruction.

\subsection{Image-Plane Spatial Decoder (IPSD)}

Without introducing additional inputs, we utilize PCE and ACM to extract distance confidence weights and geometric compensation features respectively. PCE provides pixel-level distance and position awareness based on diffuse reflection imaging. ACM enforces rigid geometric constraints through affine invariance mechanisms. However, these geometric priors remain encoded as 2D feature weights, requiring effective integration into 3D spatial representation.

To achieve theoretically grounded fusion of multi-source geometric priors, we design the Image-Plane Spatial Decoder (IPSD). IPSD introduces a theoretical framework for multi-source geometric prior fusion. We formulate cost volume construction as a joint optimization problem that simultaneously preserves distance-aware confidence from PCE, affine-invariant geometric structures from ACM, and multi-view consistency constraints. This formulation fundamentally redefines the back-projection paradigm from simple ray aggregation to geometry-aware spatial encoding, addressing the long-standing information bottleneck in sparse-view reconstruction. Specifically, IPSD establishes a nonlinear mapping from 2D features to 3D space through linear attention mechanisms. Each voxel's occupancy probability depends not only on multi-view projections but also on modulation by intra-view geometric priors. Through this design, IPSD achieves the first joint modeling of diffuse reflection imaging principles and affine invariance, providing a theoretical foundation for high-fidelity reconstruction under sparse views.

For features $CR_{i}^{r} \in \mathbb{R}^{B \times C_{cr} \times H \times W}$ that integrate multi-source geometric priors (containing distance and position information from PCE and rigid geometric constraints from ACM), we propose a state-space-based geometry-aware projection strategy. We first serialize features into $\widehat{CR}_{i}^{r} \in \mathbb{R}^{B \times C_{cr} \times T}$, then project them into state space to compute geometric correlations:

\begin{equation}
\text{State}_{i}^{r} = F_{project}(\widehat{CR}_{i}^{r}) \cdot \mathcal{A} + \mathcal{B}_{i}.
\end{equation}

Here, $F_{project}(\cdot)$ denotes state-space projection, $\mathcal{A}$ represents distance-aware confidence weights, and $\mathcal{B}_{i}$ is dynamic bias. The key innovation lies in state-space modeling, which effectively captures intrinsic correlations among distance, position, and geometric constraints rather than simple feature concatenation. The recursive nature of state space enables modeling long-range geometric dependencies, fundamentally enhancing cost volume representation capability. We incorporate preliminary back-projection features $F_{BP}^{r}$ as residual connections to preserve multi-view information. Finally, through spatial interpolation, we obtain the enhanced cost volume $TCV_{i}^{r}$, achieving deep fusion of intra-view geometric priors and multi-view constraints:

\begin{equation}
T C V_{i}^{r}=MLP_{D}^{r}\left[F_{spatial }\left(\text{State}_{i}^{r}\right), F_{BP}^{r}\right].
\end{equation}

In this formulation, $F_{{spatial }}(\cdot)$ represents spatial neighbor interpolation. Through synergistic integration of PCE, ACM, and IPSD, we achieve theory-driven multi-source geometric prior fusion. PCE extracts distance and position awareness, ACM enforces rigid geometric constraints, and IPSD establishes intrinsic correlations among these priors via state-space modeling. This systematic geometric mining mechanism enables single views to provide constraints traditionally requiring multiple views. IPDRecon fundamentally transforms the sparse-view reconstruction paradigm—from multi-view ray aggregation to joint optimization of intra-view geometric mining and multi-view constraints.

\subsection{Loss Function}

The supervision method employs a three-stage approach: view fusion loss, occupancy prediction loss, and TSDF volume regression loss. In the multi-view fusion phase, we use the cross-entropy loss as ${L}_{{F}}^{r}$ to supervise the fusion weights $W^{r}$. For occupancy prediction $O^{k}$, we first compute the occupancy state $S_{{state }}^{k}={ Argmax }\left(P_{O}^{k}\right)$, where $P_{O}^{k} = Softmax\left({O}^{k}\right)$ and use the binary cross-entropy loss as ${L}_{p}^{k}$ for supervision.

Here, $N^{k}$ denotes the total number of voxels at resolution $k\in\{ coarse,medium \}$, and $GT^{r}$ represents the occupancy ground truth at resolution. Finally, we supervise the TSDF volume regression using L1-loss. The overall loss is the sum of the three-stage losses:

\begin{equation}
{L}=\sum{L}_{{F}}^{r }+\sum{L}_{p}^{k}+{L}_{{TSDF}}^{fine }.
\end{equation}

\section{Experiment}

In this section, we provide detailed testing conditions. Subsequently, we present the test results of IPDRecon on three mainstream datasets. Additionally, the ablation experiments also demonstrate the effectiveness of our design.

\begin{table}[t]
\centering
\caption{Definitions of 2D and 3D metrics. Where, $n$ denotes the number of pixels with both valid ground truth and prediction, $d$ and $d^{gt}$ denote the predicted and the ground-truth depths, $p$ and $p^{gt}$ denote the predicted and the ground-truth point clouds.}
\begin{tabular}{c|c}
    \hline
    Metrics & Definition\\
    \hline
    Abs Rel & $\frac{1}{n} \sum\left|d-d^{gt}\right| / d^{gt}$ \\
    Sq Rel & $\frac{1}{n} \sum\left|d-d^{gt}\right|^{2} / d^{gt}$  \\
    RMSE & $\sqrt{\frac{1}{n} \sum\left|d-d^{gt}\right|^{2}}$  \\
    $\delta_{1.25}$ & $\frac{1}{n} \sum\left(\max \left(\frac{d}{d^{gt}}, \frac{d^{gt}}{d}\right)<1.25^{i}\right)$ \\
    \hline
    Acc & ${mean}_{p \in P}\left(\min _{p^{gt} \in P^{gt}}\left\|p-p^{gt}\right\|\right)$  \\
    Comp & ${mean}_{p^{gt} \in P^{gt}}\left(\min _{p \in P}\left\|p-p^{gt}\right\|\right)$  \\
    Chamf. & $\frac{{ Acc }+ { Comp }}{2}$  \\
    Prec & ${mean}_{p \in P}\left(\min _{p^{gt} \in P^{gt}}\left\|p-p^{gt}\right\|<0.05\right)$ \\
    Recall & ${mean}_{p^{gt} \in P^{gt}}\left(\min _{p \in P}\left\|p-p^{gt}\right\|<0.05\right)$  \\
    F-score & $\frac{2 \times { Prec } \times { Recall }}{{ Prec }+ { Recall }}$  \\
    \hline
\end{tabular}
\label{table2}
\end{table}

\subsection{Implementation details}
IPDRecon is built on the Pytorch \citep{pytorch} deep learning toolchain. We train IPDRecon with a single Nvidia A40 GPU, using the Adam optimizer for network optimization. Hyperparameters are set accordingly $\beta_{1}=0.9$, $\beta_{2}=0.999$, ${lr}=1 \times 10^{-3}$, with the learning rate during the fine-tuning phase set with ${lr}=1 \times 10^{-4}$. The voxel sizes are set from coarse to fine resolutions at $16cm$, $8cm$, and $4cm$, respectively. The training process is constrained to 20 views, while the testing phase is set to 100 views. 

\subsection{Datasets}
We train and test IPDRecon based on ScanNet V2 \citep{scannet} datasets. Additionally, to demonstrate the generalization performance of IPDRecon, we select 13 scenes from the TUM-RGBD \citep{tum} dataset and 8 scenes from the ICL-NUIM \citep{iclu} dataset for testing.

\subsection{Metrics}
We use the 3D and 2D metrics defined by Atlas \citep{atlas} for evaluation. We conduct our tests using the codes provided by \citep{vortx}. Furthermore, we employ the protocol provided by \citep{transformerfusion} to mitigate the penalization for occluded areas. The specific definitions of evaluation metrics are presented in Table \ref{table2}.

\begin{table}[t]
\centering
\caption{Evaluation of the 3D meshes on ScanNet. The upper part follows the evaluation protocol in \citep{vortx} and the lower part follows the evaluation protocol in \citep{transformerfusion}.}
\resizebox{\columnwidth}{!}{
\begin{tabular}{c|c|cccccc}
\hline
Methods & Core Idea & Acc $\downarrow$ & Comp $\downarrow$ & Chamf. $\downarrow$ & Prec $\uparrow$ & Recall $\uparrow$ & F-score $\uparrow$ \\
\hline
Atlas\citep{atlas} & Real-time Reconstruction & 0.068 & 0.098 & 0.083 & 0.640 & 0.539 & 0.583 \\
DeepVideoMVS\citep{deepvideomvs} & Fusion MVS Baseline & 0.079 & 0.133 & 0.106 & 0.521 & 0.454 & 0.474 \\
NeuralRecon \citep{neuralrecon} & Real-time Reconstruction & 0.054 & 0.128 & 0.091 & 0.684 & 0.479 & 0.562 \\
VoRTX\citep{vortx} & Multi-resolution Baseline & 0.054 & 0.090 & 0.072 & 0.708 & 0.588 & 0.641 \\
IOAR\citep{inner} & Occupy Supervision & \textbf{0.043} & 0.090 & 0.067 & 0.748 & 0.597 & 0.663 \\
\hline
Ours & Geometry-aware BP & 0.046 & \textbf{0.083} & \textbf{0.065} & \textbf{0.751} & \textbf{0.613} & \textbf{0.675} \\
\hline
COLMAP\citep{pixelwise1} & Incremental Reconstruction & 0.102 & 0.119 & 0.111 & 0.509 & 0.474 & 0.489 \\
MVDepthNet\citep{mvdepthnet} & Multi-view‌ Depth Fusion & 0.129 & 0.083 & 0.106 & 0.443 & 0.487 & 0.460 \\
GP-MVS\citep{dp} & Plane-sweep Cost Volume & 0.129 & 0.080 & 0.105 & 0.453 & 0.510 & 0.477 \\
DPSNet\citep{dpsnet} & Plane-sweep Cost Volume & 0.119 & 0.076 & 0.098 & 0.474 & 0.519 & 0.492 \\
Atlas\citep{atlas} & Real-time Reconstruction & 0.072 & 0.076 & 0.074 & 0.675 & 0.605 & 0.636 \\
ESTDepth\citep{est} & Semantic-Enhanced Depth & 0.127 & 0.075 & 0.101 & 0.456 & 0.542 & 0.491 \\
DeepVideoMVS\citep{deepvideomvs} & Fusion MVS Baseline & 0.107 & 0.069 & 0.087 & 0.541 & 0.592 & 0.563 \\
NeuralRecon\citep{neuralrecon} & Real-time Reconstruction & 0.051 & 0.091 & 0.071 & 0.630 & 0.612 & 0.619 \\
TransformerFusion\citep{transformerfusion} & Multi-view Fusion Baseline & 0.055 & 0.083 & 0.064 & 0.728 & 0.600 & 0.655 \\
VoRTX\citep{vortx} & Multi-resolution Baseline & 0.043 & 0.072 & 0.078 & 0.767 & 0.651 & 0.703 \\
SimpleRecon\citep{simplerecon} & 4D Metadata Cost Volume & 0.055 & \textbf{0.061} & 0.058 & 0.686 & 0.658 & 0.671 \\
SDFFormer\citep{SDFFORMER}$\dagger$ & Transformer 3D Backbone & 0.048 & 0.065 & 0.057 & 0.751 & 0.665 & 0.703 \\
IOAR\citep{inner} & Occupy Supervision & 0.037 & 0.072 & 0.055 & 0.791 & 0.650 & 0.712 \\ 
EPRecon\citep{eprecon} & Efficient Real-time Framework & 0.051 & 0.101 & 0.076 & 0.692 & 0.519 & 0.593 \\ 
SDFUtrans\citep{li2025hybrid} & Hybrid 3D Backbone & 0.045 & 0.066 & 0.056 & 0.767 & \textbf{0.671} & 0.714\\ 
DetailRecon\citep{detailrecon}$\dagger$ & Global-structural Detail & 0.037 & 0.084 & 0.061 & 0.791 & 0.601 & 0.684\\ 
\hline
Ours & Geometry-aware BP & \textbf{0.036} & 0.070 & \textbf{0.053} & \textbf{0.797} & 0.660 & \textbf{0.722} \\
\hline
\end{tabular}
}
\label{table3}
\end{table}

\begin{table}[t]
\centering
\caption{Evaluation of the 2D depth map on ScanNet.}
\small
\begin{tabular}{c|cccc}
\hline
Methods & Abs Rel $\downarrow$ & Sq Rel $\downarrow$ & RMSE $\downarrow$ & $\delta_{1.25}$ $\uparrow$ \\
\hline
COLMAP\citep{pixelwise} & 0.137  & 0.138 & 0.502 & 0.834 \\
MVDepthNet\citep{mvdepthnet} & 0.098  & 0.061 & 0.293 & 0.896 \\
GP-MVS\citep{dp} & 0.130  & 0.339 & 0.472 & 0.906 \\
DPSNet\citep{dpsnet} & 0.087  & 0.035 & 0.232 & 0.925 \\
Atlas\citep{atlas} & 0.065  & 0.043 & 0.251 & 0.936 \\
NeuralRecon\citep{neuralrecon} & 0.065  & \textbf{0.031} & 0.195 & 0.948 \\
VoRTX\citep{vortx} & 0.061  & 0.038 & 0.205 & 0.943 \\
IOAR\citep{inner} & 0.052  & \textbf{0.031} & 0.182 &  0.958 \\
SDFUtrans\citep{li2025hybrid} & 0.055  & 0.035 & 0.202 & 0.954 \\
\hline
Ours & \textbf{0.051} & \textbf{0.031} & \textbf{0.179} & \textbf{0.959} \\
\hline
\end{tabular}

\label{table4}
\end{table}

\subsection{Evaluation}
\textbf{Quantitive Analysis.} To evaluate the performance of IPDRecon, we utilized two protocols for 3D surface reconstruction quality assessment. Additionally, we compared our method with the state-of-the-art methods available on ScanNet, a large-scale indoor scene reconstruction dataset. As noted in previous research, Precision, Recall and F-score are considered reliable indicators of reconstruction quality \citep{atlas}. The evaluation results are reported in Table \ref{table3}. The assessment based on the VoRTX protocol \citep{vortx} shows that our model achieved the state-of-the-art results in most metrics, improving the surface reconstruction F-score from 0.663 to 0.675 and the Recall from $59.7\%$ to $61.3\%$. Similarly, the evaluation based on the second protocol \citep{transformerfusion} demonstrates that our model enhanced the surface reconstruction F-score from 0.714 to 0.722 and the Precision from $79.1\%$ to $79.7\%$. Our improvements aim to enrich the theoretical framework of the back-projection stage, addressing view quantity dependency challenges arising from reliance on idealized theories. The 3D evaluation results demonstrate the effectiveness of our design. The introduction of encoding-decoding processes based on imaging principles enhances the surface reconstruction capability of the IPDRecon model. Our model is able to make higher-quality predictions compared to existing models.

Existing volume-based methods directly predict TSDF volumes instead of depth maps. However, we still follow previous work to evaluate 2D depth maps to demonstrate the effectiveness of our proposed improvements. Following the method of \citep{atlas}, we render 3D shapes into the image plane to generate depth maps, with results detailed in Table \ref{table4}. The evaluation metrics are comparable to state-of-the-art methods. Since our model employs the same supervision strategy as previous methods, it remains prone to ambiguous depth range estimation in indoor spatial regions. Consequently, IPDRecon demonstrates modest improvements in 2D depth assessment metrics.

\begin{table*}[t]
\centering
\caption{Impact of View Numbers on Reconstruction Quality. Unlike existing methods whose reconstruction quality (precision, recall, and F-score) improves with increasing view numbers, our method demonstrates more stable performance. This indicates that our focus on exploiting hard geometric constraints within individual views effectively reduces the dependency on weak geometric constraints across views.}
\resizebox{\columnwidth}{!}{
\begin{tabular}{c|c|cccc|ccccc}
\hline
Methods & View & Chamf. $\downarrow$ & Prec $\uparrow$ & Recall $\uparrow$ & F-score $\uparrow$ & CV $\downarrow$ & F-score PRR $\uparrow$ & mean PRR $\uparrow$ & Max Drop $\downarrow$ & SI $\uparrow$\\
\hline
\multirow{3}*{VoRTX\citep{vortx}} & 60 & 0.061 & 0.752 & 0.631 & 0.685 \\
 & 80  & 0.059 & 0.763 & 0.639 & 0.694  & $1.30\%$ & $97.4\%$ & $97.5\%$ & $2.56\%$ & 0.926\\
 & 100 & 0.058 & 0.767 & 0.651 & 0.703 \\
\hline
\multirow{3}*{IOAR\citep{inner}} & 60 & 0.056 & 0.782 & 0.641 & 0.704 \\  
 & 80 & 0.055 & 0.791 & 0.649 & 0.712 & $1.05\%$ & $97.9\%$ & $98.0\%$ & $2.09\%$ & 0.943 \\ 
 & 100 & 0.054 & 0.794 & 0.657 & 0.719 \\ 
\hline
\multirow{3}*{SDFUtrans\citep{li2025hybrid}} & 60 & 0.064 & 0.758 & 0.656 & 0.703  \\   
 & 80 & 0.059 & 0.761 & 0.659 & 0.706  & 0.81\% & 98.5\% & 98.3\% & 1.54\% & 0.965\\  
 & 100 & 0.056 & 0.767 & \textbf{0.671} & 0.714 \\ 
\hline
\multirow{3}*{Ours} & 60 & 0.054 & 0.795 & 0.659 & 0.719 \\
 & 80 & 0.054 & \textbf{0.797} & 0.662 & \textbf{0.723} & \textbf{0.24\%} & \textbf{99.6\%} & \textbf{99.7\%} & \textbf{0.42\%} & \textbf{0.991} \\
 & 100 & \textbf{0.053} & \textbf{0.797} & 0.660 & 0.722 \\
\hline
\end{tabular}
}
\label{table5}
\end{table*}

\textbf{Analysis of View Quantity Impact.} As shown in Table \ref{table5}, we evaluated the impact of view quantity on reconstruction quality through comprehensive stability metrics. To quantitatively assess view-invariance properties, we introduce four stability indicators: Coefficient of Variation (CV), Performance Retention Rate (PRR), Maximum Performance Drop (Max Drop), and Stability Index (SI).

(1) Coefficient of Variation (CV): This metric measures relative variability, defined as 

\begin{equation}
CV = \sigma/\mu \times 100\%,
\end{equation}

where $\sigma$ is the standard deviation and $\mu$ is the mean across different view numbers. Lower CV indicates more stable performance.

(2) Performance Retention Rate (PRR): This metric quantifies performance preservation when reducing views, calculated as 
\begin{equation}
PRR = P_{60}/P_{100} \times 100\%, 
\end{equation}

where $P_{60}$ and $P_{100}$ represent performance at 60 and 100 views, respectively. We also report Mean PRR as the average retention rate across Precision, Recall, and F-score metrics.

(3) Maximum Performance Drop (Max Drop): This measures the largest performance degradation, defined as 
\begin{equation}
\text{Max Drop} = (P_{100} - P_{60})/P_{100} \times 100\%,
\end{equation}

indicating the relative performance loss under sparse views.

(4) Stability Index (SI): A comprehensive metric combining multiple stability aspects, calculated as 
\begin{equation}
SI = (1 - \Delta_F/F_{max}) \times (1 - \Delta_P/P_{max}) \times (1 - \Delta_R/R_{max}),
\end{equation}

where $\Delta$ represents the range between maximum and minimum values for F-score, Precision, and Recall respectively.

IPDRecon achieves remarkable stability with a CV of only 0.24\%, compared to 1.30\% for VoRTX, 1.05\% for IOAR, and 0.81\% for SDFUtrans, indicating 5.4$\times$, 4.4$\times$, and 3.4$\times$ better consistency, respectively. The F-score PRR of 99.6\% demonstrates that our method retains nearly identical reconstruction quality even with 40\% fewer views (60 vs. 100). The mean PRR across all metrics reaches 99.7\%, significantly outperforming VoRTX (97.5\%), IOAR (98.0\%), and SDFUtrans (98.3\%). Furthermore, the Max Drop of only 0.42\% compared to 2.56\% for VoRTX represents a 6$\times$ improvement in robustness. The comprehensive SI score of 0.987 versus 0.925 (VoRTX), 0.941 (IOAR), and 0.952 (SDFUtrans) confirms the superior stability of our approach.

These quantitative results validate that by leveraging PCE and ACM to extract rich intra-view geometric priors---including distance, position, and affine-invariant features---IPDRecon fundamentally reduces dependency on view quantity, addressing a critical limitation of existing multi-view reconstruction methods.

\begin{table}[!htbp]
\centering
\caption{Computational efficiency evaluation. We measured our model's computational performance using a single RTX 3090 GPU. The timing analysis was conducted at two distinct levels: individual frame processing and complete scene reconstruction. Frame-level assessment focused on two-dimensional feature detection processes, whereas scene-level evaluation incorporated projection reversal operations, view integration procedures, and volumetric construction techniques.}\label{tbl6}
\begin{tabular}{c|ccc}
\hline
Methods & Per Frame Time & Per Scene Time & Time\\
\hline
Atlas\citep{atlas} & 71ms & 840ms & 14\\
NeuralRecon\citep{neuralrecon} & 30ms & 0ms & 33\\
TF\citep{transformerfusion} & 130.5ms & 243.3ms & 7\\
VoRTX\citep{vortx} & 71.4ms & 231.7ms & 14\\
SDFFormer\citep{SDFFORMER} & 13.3ms & 286.4ms & 75\\
IOAR\citep{inner} & 12.9ms & 254.9ms & 78 \\
SDFUtrans\citep{li2025hybrid} & 13.0ms & 262.8ms & 76\\
\hline
ours & 13.0ms & 278.5ms & 76\\
\hline
\end{tabular}
\label{table6}
\end{table}

\begin{table}[t]
\centering
\caption{Evaluation of the generalization. All models are trained on ScanNet datasets and tested on ICL-NUIM and TUM-RGBD.}
\begin{tabular}{c|c|ccc}
\hline
Datasets & Methods & Prec $\uparrow$ & Recall $\uparrow$ & F-score $\uparrow$ \\
\hline
\multirow{6}*{\rotatebox{90}{ICL-NUIM}} & Atlas\citep{atlas} & 0.280 & 0.194 & 0.229 \\
 & NeuralRecon\citep{neuralrecon} & 0.214 & 0.036 & 0.058 \\
 & VoRTX\citep{vortx} & 0.449 & 0.375 & 0.408 \\
 & SDFFormer$\dagger$\citep{SDFFORMER} & 0.522 & 0.390 & 0.447\\
 & IOAR$\dagger$\citep{inner} & 0.542 & 0.363 & 0.433 \\
 & SDFUtrans\citep{li2025hybrid} & 0.527 & \textbf{0.401} & \textbf{0.455} \\
\hline
 & Ours & \textbf{0.544} & 0.383 & 0.449 \\
\hline
\multirow{6}*{\rotatebox{90}{TUM-RGBD}} & Atlas\citep{atlas} & 0.360 & 0.089 & 0.132 \\
 & NeuralRecon\citep{neuralrecon} & 0.382 & 0.075 & 0.115 \\
 & VoRTX\citep{vortx} & 0.280 & 0.194 & 0.229 \\
  & SDFFormer$\dagger$\citep{SDFFORMER} & 0.406 & 0.194 & 0.254\\
 & IOAR$\dagger$\citep{inner} & 0.540 & 0.240 & 0.332 \\
 & SDFUtrans\citep{li2025hybrid} & 0.372& 0.206 & 0.265\\
\hline
 & Ours & \textbf{0.541} & \textbf{0.260} & \textbf{0.350} \\
\hline
\end{tabular}

\label{table7}
\end{table}

\textbf{Efficiency Analysis.} Computational efficiency metrics are displayed in Table \ref{table6}. We adopted the evaluation protocol established by SDFFormer \citep{SDFFORMER} and executed all tests utilizing a singular NVIDIA RTX 3090 GPU. Voxel chunks were configured at 1.5×1.5×1.5m³ dimensions. Our assessment framework incorporates two distinct evaluation phases: initially, frame-level analysis encompassing 2D feature detection processes, followed by scene-level evaluation that integrates back-projection operations, view synthesis consolidation, and volumetric reconstruction procedures. The frame-level assessment exclusively utilizes the 2D convolutional neural architecture backbone, thus performance is fundamentally constrained by the feature extraction mechanism. Despite implementing various optimization techniques, enhancement opportunities remained constrained in this domain. IPDRecon incorporated a novel linear attention mechanism to construct back-projection. Due to suboptimal memory utilization of this inference mechanism, processing speed was significantly affected. Nevertheless, we still maintained competitive inference performance.

\textbf{Generalization Evaluation.} To validate the cross-domain generalization capability of our model, we adopt a protocol of training on ScanNet and testing on ICL-NUIM and TUM-RGBD datasets. As shown in Table \ref{table7}, IPDRecon achieves performance improvements on both test datasets: precision increases from 0.542 to 0.544 on ICL-NUIM, while recall and F-score improve from 0.240 and 0.332 to 0.260 and 0.350 respectively on TUM-RGBD. The above generalization performance improvements validate IPDRecon's cross-scene adaptability. We attribute this primarily to the design philosophy of IPDRecon components that exploit spatial geometric information through pixel correlations within the image plane. Compared to traditional methods that rely on specific data distributions, our approach emphasizes the physical essence of geometric optical imaging, thereby achieving stronger scene-agnostic properties and generalization potential.

\begin{figure*}[t]
\centering
\includegraphics[width=\textwidth]{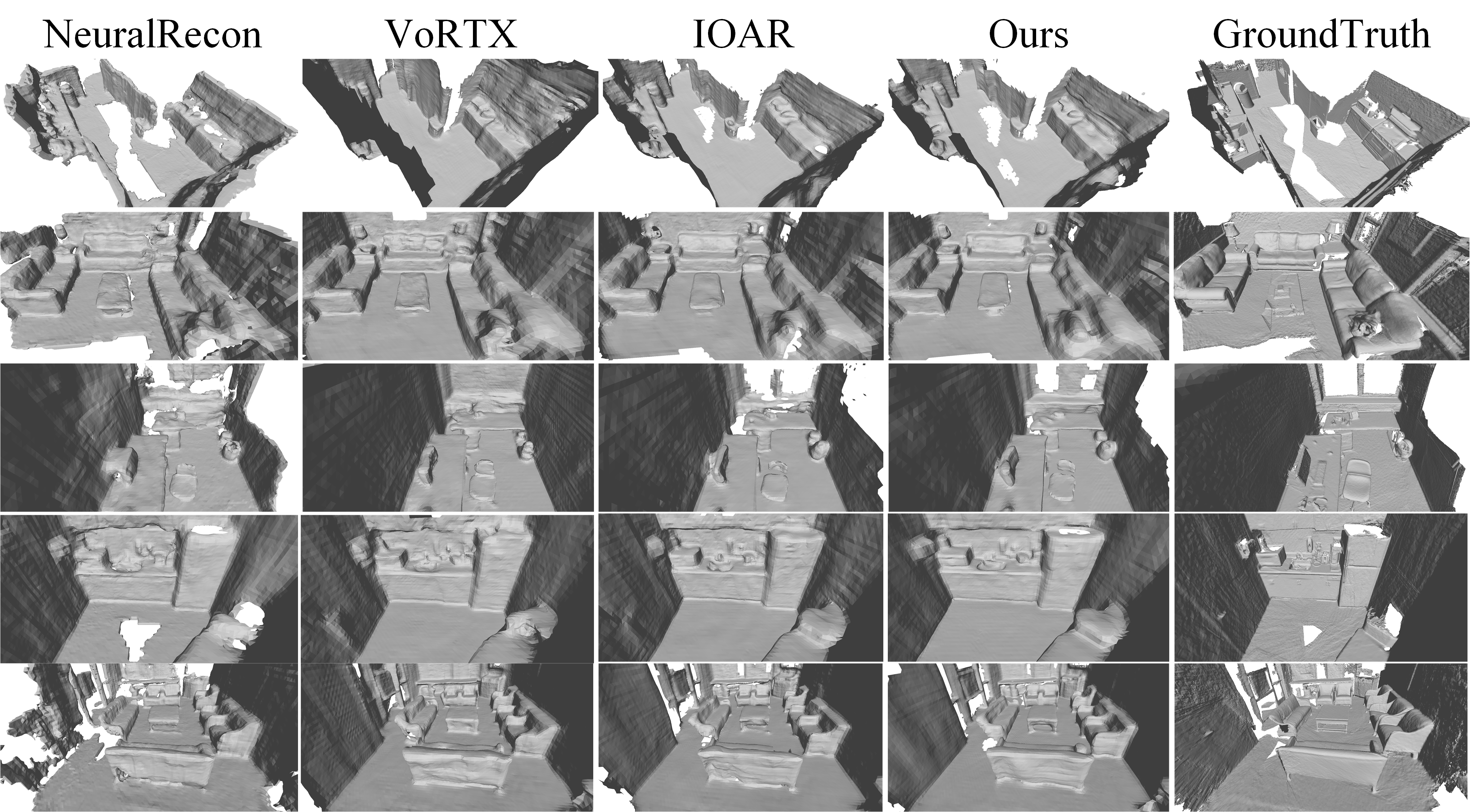} 
\caption{Qualitative analysis of representative scenarios. Comparative evaluation against three conventional approaches demonstrates that our method achieves superior reconstruction fidelity.}
\label{fig7}
\end{figure*}

\begin{figure*}[t]
\centering
\includegraphics[width=\textwidth]{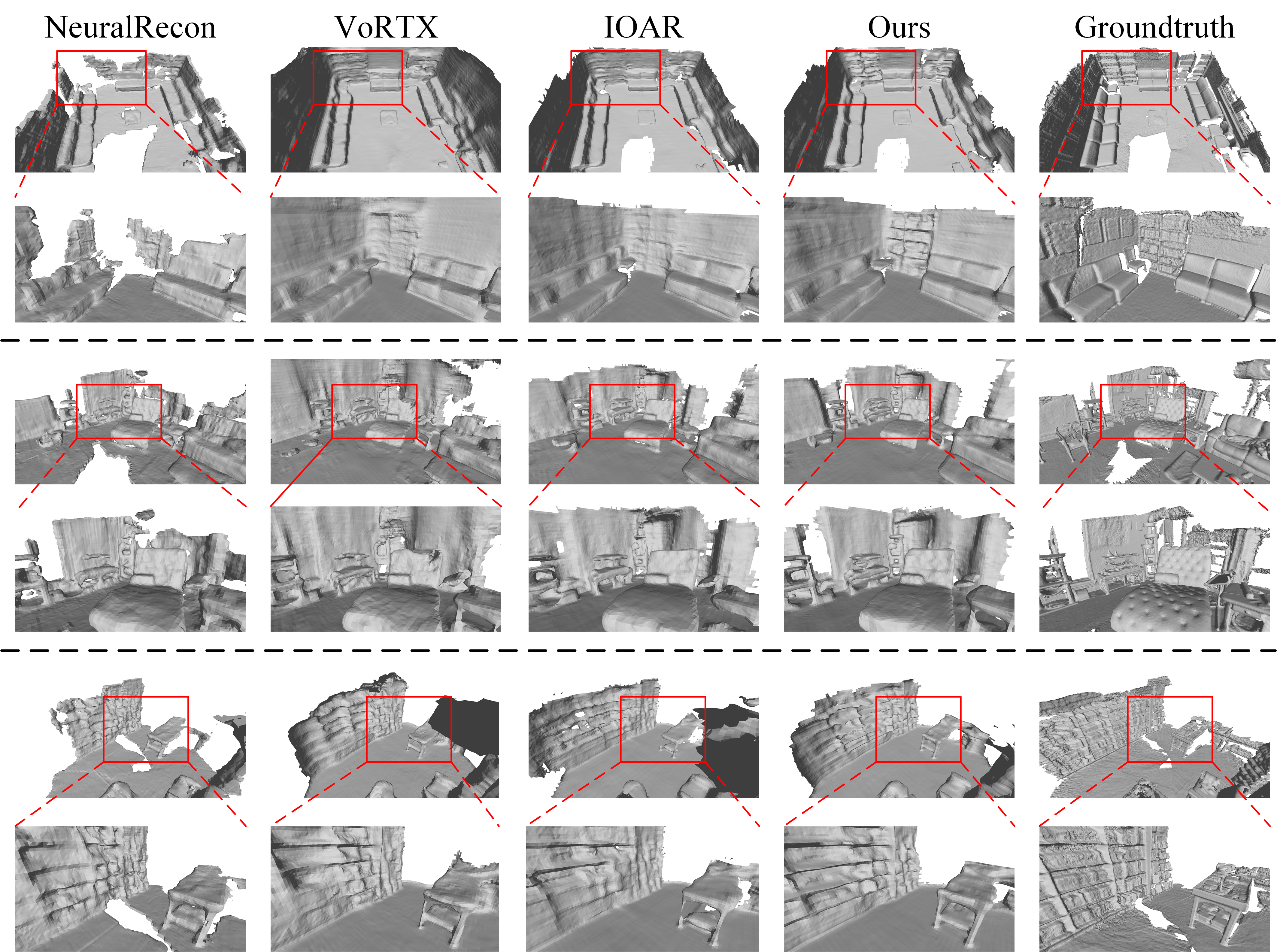} 
\caption{Detailed Examination of Qualitative Analysis. From left to right are the surface reconstructed by NeuralRecon, VoRTX, IOAR and our model. And the most right column contains the ground-truth meshes. The red boxes in the figure highlight some detailed visual results. These details provide a clearer illustration of our model's contributions. }
\label{fig8}
\end{figure*}

\textbf{Qualitative Analysis.} To more intuitively demonstrate the performance of IPDRecon, we conducted a 3D surface qualitative analysis and compared it with existing methods. The visualization results are shown in Figure \ref{fig7} and Figure \ref{fig8}. By systematically exploiting intra-view geometric information through PCE, ACM, and IPSD, our model reconstructs higher quality surfaces with enhanced precision and completeness, particularly in sparse viewpoint scenarios. Additionally, as highlighted in the red boxes in Figure \ref{fig8}, our model produces superior predictions for hollow and thin structures by leveraging hard geometric constraints from affine transformation principles. For instance, in the third scene, our model reconstructs a sofa model with richly detailed textures by mining geometric information encoded within individual views, whereas existing methods relying solely on multi-view ray intersections produce overly smooth surfaces with insufficient detail processing. The superior performance in reconstructing detailed features such as table lamps, sofas, and desktops demonstrates that IPDRecon effectively addresses the fundamental limitation of weak geometric constraints in existing frameworks, achieving high-fidelity reconstruction even with limited view inputs.

\begin{figure*}[t]
\centering
\includegraphics[width=\textwidth]{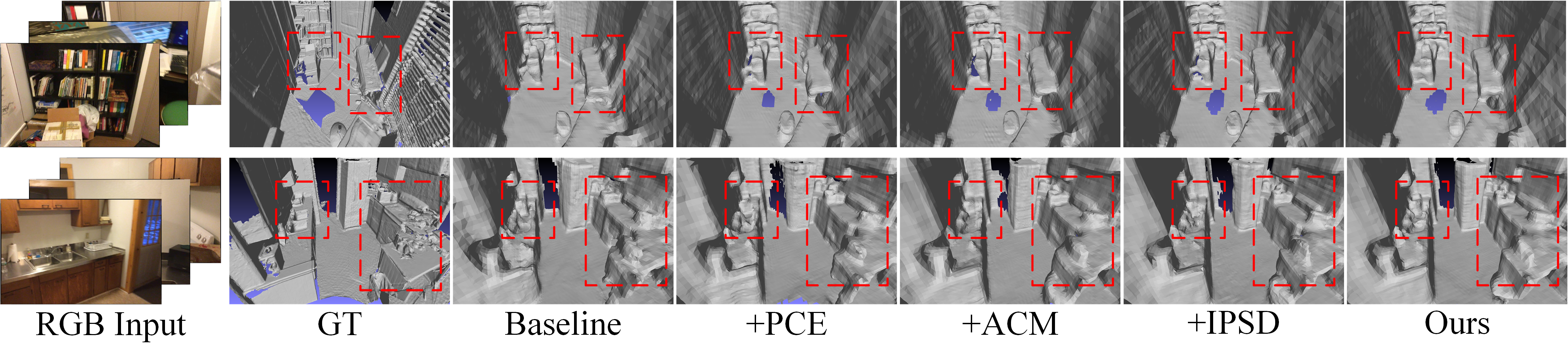} 
\caption{Qualitative Analysis of Ablation Experiments. We demonstrate the individual impact of adding PCE, ACM, and IPSD modules on reconstruction quality. Compared to the baseline, our proposed modules achieve superior performance in spatial detail preservation and noise suppression. Specifically, ACM produces more conservative predictions due to the incorporation of hard geometric constraints. The joint utilization of all three modules enables our model to exhibit robust performance.}
\label{fig9}
\end{figure*}

\begin{table*}[t]
\centering
\caption{Evaluation of ablation study. We conducted eight setups of ablation experiments. The first setup represents the baseline pipeline. The second to eighth setups examine the individual and combined effects of various components of IPDRecon.}
\resizebox{\columnwidth}{!}{
\begin{tabular}{c|c|ccc|cccccc}
\hline
Num. & Settings & PCE & ACM & IPSD & Acc $\downarrow$ & Comp $\downarrow$ & Chamf. $\downarrow$ & Prec $\uparrow$ & Recall $\uparrow$ & F-score $\uparrow$ \\
\hline
1 & Baseline & - & - & - & 0.038 & 0.079 & 0.059 & 0.762 & 0.637 & 0.694 \\
\hline
2 & \multirow{3}*{Single Component} & \checkmark & - & - & 0.040 & 0.076 & 0.058 & 0.778 & 0.642 & 0.702 \\
3 &  & - & \checkmark & - & 0.038 & 0.080 & 0.059 & 0.776 & 0.636 & 0.698 \\
4 &  & - & - & \checkmark & 0.040 & 0.078 & 0.059 & 0.777 & 0.635 & 0.697 \\
\hline
5 & \multirow{3}*{Double Components} & \checkmark & \checkmark & - & 0.040 & 0.077 & 0.059 & 0.782 & 0.642 & 0.704 \\
6 &  & - & \checkmark & \checkmark & 0.039 & 0.077 & 0.058 & 0.780 & 0.636 & 0.699 \\
7 &  & \checkmark & - & \checkmark & 0.040 & 0.081 & 0.061 & 0.775 & 0.623 & 0.689 \\
\hline
8 & Components Fusion & \checkmark & \checkmark & \checkmark & \textbf{0.036} & \textbf{0.070} & \textbf{0.053} & \textbf{0.797} & \textbf{0.660} & \textbf{0.722} \\
\hline
\end{tabular}
}
\label{table8}
\end{table*}

\subsection{Ablation Study}
To further explore the contributions of various optimizations and components in IPDRecon, we conducted an ablation study. The results are reported in Table \ref{table8}. We designed eight different experimental setups for thorough evaluation. The first setup represents the basic pipeline. In the subsequent second through eighth sets of experiments, we tested our three novel design components both individually and in various combinations within different framework configurations.

\textbf{Independent framework implementation.} The second through fourth sets demonstrate the test results of independently applying PCE, ACM, and IPSD within the optimized pipeline. When employed individually, PCE and IPSD function primarily as attention layers. Through linear attention mechanisms, they assign attention weights to each pixel within the plane. During training, these attention weights enhance focus on surfaces and edges. Consequently, the introduction of PCE and IPSD improves three-dimensional reconstruction quality. Evaluation results demonstrate that incorporating PCE and IPSD increases precision from $76.3\%$ to $77.8\%$ and $77.7\%$, while raising F-score from 0.694 to 0.702 and 0.697, respectively. Furthermore, although IPDRecon does not introduce additional views, ACM effectively incorporates features with rich three-dimensional geometric properties into the network. Consequently, the introduction of ACM similarly enhances reconstruction quality, improving precision and F-score to $77.6\%$ and 0.698.

\textbf{Combined framework implementation.} The fifth through seventh sets demonstrate the evaluation results of pairwise combinations within the optimized pipeline. First, PCE provides preliminary pixel plane encoding information. ACM incorporates three-dimensional spatial information based on established distance and position information. This combination helps the network enhance reconstruction quality. Experimental results show precision and F-score improving to $78.2\%$ and 0.704. Furthermore, decoding after enriching spatial information through ACM also improves performance. Experimental results show precision and F-score increasing to $78.0\%$ and 0.699. However, employing solely encoding and decoding processes without supplementary spatial information fails to enhance reconstruction quality. Immediate decoding following encoding generates redundant processes, hindering reconstruction and elevating computational complexity. Consequently, precision and F-score decline to $77.5\%$ and 0.689, inferior to metrics with independent PCE and IPSD implementation. 

However, neither individual additions nor pairwise combinations achieve optimal results. IPDRecon functions more effectively as an integrated system. First, PCE encodes each pixel in the image plane, establishing fundamental distance and positional relationships. Subsequently, building on PCE encoding, ACM incorporates hard constraints with rich geometric priors without introducing additional views. Finally, IPSD decodes the distance, position, and spatial geometric information embedded within pixel-level weights, significantly enhancing reconstruction quality. Compared to the baseline, the joint implementation of PCE, ACM, and IPSD improves precision and F-score by $5.98\%$ and $5.40\%$.

\textbf{Qualitative Analysis of Ablation Experiments.} To further demonstrate the impact of each IPDRecon component on performance, we provide visualization results of reconstructions with individual component additions, as shown in Figure \ref{fig9}. Spatial details of interest are highlighted by red dashed boxes in the figure. PCE decodes distance and position information as geometric constraints based on diffuse imaging modeling. While enabling the network to acquire preliminary 3D perception capabilities, it effectively maintains spatial details and filters noise voxels. However, the addition of geometric constraints makes the network tend toward conservative predictions of voxel occupancy in space, leading to missing parts of structures. ACM further introduces hard geometric constraints by leveraging affine invariance principles. The addition of ACM makes the reconstructed geometric structures more angular to some extent, but network predictions become more conservative due to increased geometric constraints. This causes the network to be more inclined to ignore fine structural decisions. In contrast, IPSD exhibits more aggressive behavior during prediction since it only utilizes existing information. This also makes spatial noise difficult to filter effectively. Therefore, compared to using individual components separately, joint utilization fully realizes our concept of exploiting intra-view 3D information. This usage effectively balances multiple constraints including position, distance, and hard geometry, filtering spatial noise while maintaining detailed geometric structures.

\section{Conclusion}

We present IPDRecon, an image-plane decoding framework for indoor reconstruction. Addressing the fundamental limitation that existing methods rely solely on weak geometric constraints from multi-view ray intersections, we systematically exploit geometric information within individual views to enhance reconstruction stability. Through PCE and ACM, we extract distance, positional, and affine-invariant features from single views. IPSD then integrates these intra-view priors with multi-view constraints. Experiments on ScanNetV2 demonstrate that IPDRecon significantly improves reconstruction stability: achieving a coefficient of variation of 0.24\% (5.4× lower than VoRTX), performance retention rate of 99.7\%, and maximum performance drop of only 0.42\% when reducing views by 40\%. IPDRecon achieves 79.7\% precision and 0.722 F-score under standard evaluation conditions, providing a robust reconstruction solution for view-limited scenarios.

\textbf{Limitations.} While IPDRecon significantly reduces dependency on view quantity and enhances reconstruction quality, several limitations remain. Due to sparse structure constraints, occupancy prediction may negatively impact reconstruction of fine details such as perforations and support structures. Additionally, PCE based on diffuse reflection imaging principles may face challenges when processing extreme lighting conditions and highly reflective surfaces. Meanwhile, affine transformation assumptions in ACM may be insufficient when handling non-rigid or highly deformed structures. In future work, we will implement further updates to the overall framework. We aim to develop more compact reconstruction architectures to overcome sparse structure constraints and explore incorporating nonlinear geometric transformations into the framework to handle more complex scene geometry.

\section*{Acknowledgement}
This work was supported partially by the Chinese Aviation Science Fund (No.20230058048013).

\bibliographystyle{elsarticle-harv} 
\bibliography{ref}
\end{document}